\documentclass[a4paper,11pt,twocolumn,twoside]{article}
\usepackage[dvips]{graphicx}
\usepackage{sepln_en}
\usepackage{fullname}
\usepackage[utf8]{inputenc}
\input epsf

\usepackage{latexsym}
\usepackage[T1]{fontenc}
\usepackage{lmodern}
\usepackage[utf8]{inputenc}
\usepackage{microtype}
\usepackage{inconsolata}
\usepackage{graphicx}
\usepackage[disable]{todonotes}
\usepackage{float}
\usepackage{subcaption,booktabs,multirow,comment,graphicx} %
\usepackage{tabularx}
\usepackage[capitalise]{cleveref}
\usepackage{enumitem}
\usepackage{url}

\title{BERnaT: Basque Encoders for Representing \\ Natural Textual Diversity}

\author {\textbf{Ekhi Azurmendi,} \textbf{Joseba Fernandez de Landa,} \textbf{Jaione Bengoetxea,} \\ \textbf{Maite Heredia,} 
\textbf{Julen Etxaniz,} \textbf{Mikel Zubillaga,} \textbf{Ander Soraluze,} \textbf{Aitor Soroa}\\
HiTZ Center - Ixa, University of the Basque Country UPV/EHU\\
ekhi.azurmendi@ehu.eus\\
}

\seplntranstitle{BERnaT: Modelos Discriminativos en Euskera para Representar la Diversidad del Lenguaje Natural}

\seplnclave{Entrenamiento, Idiomas con pocos recursos, Corpus, Diversidad Lingüistica, Evaluación}

\seplnresumen{Los modelos de lenguaje dependen de enormes corpus textuales que se suelen someter a un filtrado de calidad que puede conllevar la exclusión de variedades lingüísticas no estándar, una menor robustez y el refuerzo de sesgos de representatividad. 
Los modelos deberían aspirar a la representación de todo el espectro de variación lingüística, en lugar de limitarse al texto estandarizado. Utilizamos el euskera como caso de estudio y creamos la familia de modelos discriminativos  BERnaT, entrenada con fuentes de mayor diversidad: redes sociales y textos históricos. El pre-entrenamiento de los modelos se realiza con tres combinaciones de textos: estándar, diversos y combinados. Nuestra propuesta de evaluación consiste en separar las tareas entre subconjuntos de lenguaje estándar y no estándar, para poder medir la generalización lingüística de los modelos. Los resultados demuestran que los modelos entrenados con una mayor diversidad lingüística superan consistentemente a aquellos basados solo en texto estándar. Así, es posible mejorar la generalización lingüística en todas las tareas sin perder precisión en las evaluaciones más tradicionales. Esto evidencia la importancia de la diversidad lingüística para lograr modelos más inclusivos con una mayor capacidad de generalización.
}

\seplnkey{Training, Low-Resource Languages, Corpus, Linguistic Diversity, Evaluation}

\seplnabstract{Language models depend on massive text corpora that are often filtered for quality, a process that can unintentionally exclude non-standard linguistic varieties, reduce model robustness and reinforce representational biases. In this paper, we argue that language models should aim to capture the full spectrum of language variation (dialectal, historical, informal, etc.) rather than relying solely on standardized text. Focusing on the Basque language, we construct new corpora combining standard, social media, and historical sources, and pre-train the BERnaT family of encoder-only models in three configurations: standard, diverse, and combined. We further propose an evaluation framework that separates Natural Language Understanding (NLU) tasks into standard and diverse subsets to assess linguistic generalization. Results show that models trained on both standard and diverse data consistently outperform those trained on standard corpora, improving performance across all task types without compromising standard benchmark accuracy. These findings highlight the importance of linguistic diversity in building inclusive, generalizable language models.}

\firstpageno{1}

\begin{document}

\setlength\titlebox{25cm} %

\label{firstpage} \maketitle

\section{Introduction}
\label{sec:introduction}

\todo{-tiempo verbal \\-ortografía consistente (americano o britanico)}

The development of large-scale language models requires large quantities of textual data that is often crawled from the web and subsequently filtered to ensure its quality. The amount of text that is left out as well as its characteristics will vary according to various factors, such as the target domain or language, as well as to differences in the beliefs of authors as to what constitutes noise \cite{laurencon2022,etxaniz2024latxaopenlanguagemodel}. Unwanted text may include code, text in languages other than the target language, or pure gibberish. 

In the process of corpus creation and filtering, some authors may opt to exclude language varieties that, although representative of natural language use, do not meet their standards of quality - either intentionally, or by default through the selection of a reliably high-quality subset of the data~\cite{wenzek-etal-2020-ccnet,artetxe-etal-2022-corpus-robertaeuscrawl,soldaini-etal-2024-dolma,palomar-giner-etal-2024-curated,etxaniz2024latxaopenlanguagemodel}. 
In contrast, in this paper we claim that any corpus gathered with the final objective of training a foundational model should try to capture the full diversity of language, including variation across dialects (diatopic), registers (diaphasic), social groups (diastratic), and even time (diachronic). %
Modeling this diversity can ensure that the resulting model will be robust and capable of handling a broader range of domains and language varieties. %

In practice, it has been proven that models trained with less diverse data suffer from representation biases and problems in performance \cite{blodgett-etal-2016-demographic}. Even if this is not intended, a selection of texts to train a model or a quality filter performed without proper care is likely to be skewed towards powerful groups \cite{gururangan-etal-2022-whose}. These studies are mostly centered around English and other high-resource languages.
\begin{figure}
    \centering    \includegraphics[width=0.7\linewidth]{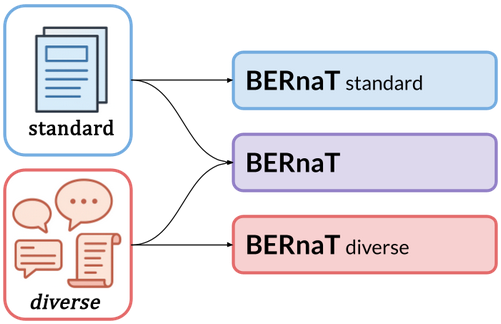}
    \caption{Summary of corpora combinations used to train BERnaT models. 
    \textit{Standard Corpus}: high-quality standard Basque from sources like News or Wikipedia. 
    \textit{Diverse Corpora}: social media and historical texts capturing informal, dialectal, and pre-standard Basque.}
    \label{fig:model_summary}
\end{figure}
Conversely, in this work we focus on Basque, a low-resource and morphologically rich language, and conduct our experiments using encoder-based architectures. Our research's aim is therefore to contribute to the body of resources for the Basque language and develop new monolingual models designed for specific Natural Language Understanding (NLU) tasks.

As shown in Figure~\ref{fig:model_summary}, we use three different corpora combinations to pre-train a collection of models. The BERnaT$_{\mathrm{standard}}$ models are pre-trained using a widely used standard corpus, whereas BERnaT$_{\mathrm{diverse}}$ models use a newly introduced diverse corpus that covers different varieties of Basque. Finally, the BERnaT models are pre-trained with both the standard and the diverse corpora.

We propose a new configuration of NLU tasks to evaluate BERnaT models, splitting existing benchmark tasks into standard and diverse subsets depending on the register and source of the annotated data, thus emulating the standard/diverse separation. %
This evaluation framework is designed to facilitate the assessment of model performance under language diversity conditions, enabling a more comprehensive understanding of how well the models generalize across different varieties and linguistic contexts.

The following are the principal findings from our experiments, which may help dataset creation and model development for other languages:
\begin{itemize}
    \item The inclusion of non-standard text in model pre-training is helpful for NLU tasks that deal with non-standard language, such as text from social media, dialectal variations and historical texts. %
    \item The performance gain of combined BERnaT models is more pronounced when the data used for fine-tuning is scarce.
    \item Including diverse text doesn't degrade the results of traditional NLU tasks written in standard varieties (BasqueGLUE, NLI, PoS).
    \item Even when including diverse corpora in training, models perform worse in diverse evaluations, showing that these tasks remain more challenging.
    \item Model size has an impact on performance, with smaller models achieving better results if specialized either on standard or diverse texts, and larger models obtaining better when they combine both. 
    
\end{itemize}

Beyond the insights previously discussed, this work contributes several resources\footnote{URL to be announced UPON acceptance} that support future research in Basque language technology: (i) the new family of BERnaT encoder models, available in three sizes and three training configurations, designed to process linguistically diverse text; (ii) the largest diverse language corpus for Basque, containing both contemporary social media and historical texts; and (iii) a comprehensive suite of benchmark datasets, comprising both established and new tasks, categorized into standard and diverse linguistic domains.

\section{Related work}

\paragraph{Approaches to language diversity.}  

The exclusion of varieties outside of the standard can be motivated by underlying language ideologies about what constitutes clean language \cite{Walsh21102021} or the convenience of a more standardized and homogeneous language in language modeling and evaluation \cite{jorgensen-etal-2015-challenges}. Nonetheless, it often leads to misrepresentation of social groups outside of the mainstream \cite{gururangan-etal-2022-whose,blodgett-etal-2020-language} and worse performance and ability to generalize to unseen language varieties \cite{markl-mcnulty-2022-language,bengoetxea-etal-2025-lost,srirag-etal-2025-evaluating}. When faced with this challenge, previous approaches may include the \textit{normalization} of text before its processing, or domain adaptation of models \cite{han-baldwin-2011-lexical-normalization,eisenstein-2013-bad,kuparinen-etal-2023-dialect,alhafni-etal-2024-exploiting}.

\paragraph{Basque language diversity and standardization.} In the process of gathering a diverse corpus, we have identified sources which include texts characterized by diatopic and diaphasic variability, that is, different dialects and registers, as well as a wide range of topics and domains. Our main sources of non-standard language include a) texts written before the development of \textit{Euskara Batua}, the standardized form of the Basque language \cite{hualde-zuazo-200-standardization-basque,Oñederra+2016+125+144}, that use non-standard dialectal orthography \cite{speeling-normalisation-basque}; and b) more recent texts gathered from social media, which include innovative spellings as well as a more conversational and relaxed register \cite{Elordui2022BasqueInstagram,fernandez-de-landa-etal-2024-uncovering}.

\paragraph{Encoder models and corpora.} While decoder-only LLMs are popular due to their text generation capabilities, encoder-only models are still frequently used for classification, clustering and retrieval, even beating decoders that are much larger~\cite{weller2025seqvsseqopen,gisserotboukhlef2025pretrainencodersmaskedlanguage}. The majority of encoder models have been trained on standard corpora, such as Wikipedia or journalistic texts, both in English \cite{liu2019roberta} and multilingual settings \cite{conneau-etal-2020-unsupervised}. However, several efforts have also focused on training encoder models on diverse corpora from social media in English \cite{nguyen-etal-2020-bertweet} and other languages \cite{barbieri-etal-2022-xlm}, and on evaluating them on that specific source \cite{barbieri-etal-2020-tweeteval}. Recently, works have centered on training encoder models with modern techniques and corpora scales closer to LLMs for English \cite{modernbert}, and the same approach has been extended to cover around 1.8K languages \cite{marone2025mmbertmodernmultilingualencoder}. However, these models have been mostly trained on standard filtered texts, with no special focus on training or evaluating on diverse corpora.

\paragraph{Basque encoder models and corpora.} Current SoTA Basque monolingual encoders such as RoBERTa EusCrawl \cite{artetxe-etal-2022-corpus-robertaeuscrawl} and ElhBERTeu \cite{urbizu-etal-2022-basqueglue} were mostly trained on standard text that mostly comes from news websites and Wikipedia. \cite{artetxe-etal-2022-corpus-robertaeuscrawl} found that using other lower-quality corpora, such as mC4 \cite{xue-etal-2021-mt5} and CC-100 \cite{conneau-etal-2020-unsupervised}, provided similar results compared to Euscrawl. Recently, bigger corpora such as Latxa corpus \cite{etxaniz2024latxaopenlanguagemodel} and ZelaiHandi \cite{ZelaiHandi} have been introduced, but no encoder models have been trained on these. These corpora, despite having more diverse sources, still lack language diversity, which is very minimal as shown in \cref{sec:corpora}. In contrast, our new corpora include a larger number of diverse texts.

\begin{table}[t] \footnotesize 
\footnotesize
  \centering
   \begin{tabular}{@{}lrrl@{}} \toprule
    Source                   & Docs  & Words & Diversity   \\ \midrule
    \textbf{Wikipedia}       & 409k  &   51M & $0.1e^{-2} \pm0.2e^{-1}$ \\
    \textbf{Egunkaria}       & 176k  &   39M & $0.3e^{-2} \pm0.2e^{-1}$ \\
    \textbf{EusCrawl-v1.1}   & 1.79M &  359M & $0.4e^{-2} \pm0.4e^{-1}$ \\
    \textbf{Colossal-oscar}  & 234k  &  105M & $1.8e^{-2} \pm0.8e^{-1}$ \\
    \textbf{CulturaX}        & 1.31M &  541M & $1.9e^{-2} \pm0.8e^{-1}$ \\
    \textbf{HPLT-v1}         & 375k  &  120M & $3.0e^{-2} \pm1.1e^{-1}$ \\
    \textbf{Booktegi}        & 166   &    3M & $7.3e^{-2} \pm1.3e^{-1}$ \\ \midrule
    \textbf{BSM} \textcolor{red}{\bf{[new]}}     &  13k  &  188M & $1.4e^{-1} \pm1.7e^{-1}$ \\  
    \textbf{EKC} \textcolor{red}{\bf{[new]}}             & 338   &   21M & $7.3e^{-1} \pm1.7e^{-1}$  \\\bottomrule
  \end{tabular}
  \caption{Corpora sizes and diversity analysis for \emph{latxa-corpus-v1.1} (upper part) and our proposed \emph{Diverse Corpora} (bottom part). Diversity is calculated per document as described in Section \ref{sec:diversity}, and we also include the standard deviation.}
  \label{tab:train-data-analysis}
\end{table}

\section{Training Corpora} 
\label{sec:corpora}

To represent diverse language usage in Basque, we combine an existing large-scale standard corpus (\emph{latxa-corpus-v1.1}) with a newly curated corpus called \emph{Diverse Corpora}, which is composed of varied and non-standard collections of texts (c.f. Section \ref{sec:diverse_corpora}). This approach allows us to capture both formal, well-edited language and informal, dialectal, or historical varieties. We perform an empirical analysis of the selected corpora, quantifying their linguistic diversity and confirming their alignment with the intended categories, therefore ensuring that our pre-training data reflects a broad spectrum of Basque language use. %

\subsection{Standard Language Corpus}
\label{sec:standard_corpora}

We utilize the largest standard Basque corpus currently available, \emph{latxa-corpus-v1.1} %
\cite{etxaniz2024latxaopenlanguagemodel}. The corpus comprises 4.3M documents in standard Basque, summing up to 1.22B words. Its sources include EusCrawl-v1.1, extracted using ad-hoc scrapers \cite{artetxe-etal-2022-corpus-robertaeuscrawl}, as well as Egunkaria, containing content from the daily Basque newspaper Euskaldunon Egunkaria (2001–2006), and Booktegi, which provides a collection of literature books in Basque. Additional sources include Wikipedia, as well as datasets derived from Common Crawl, namely CulturaX \cite{nguyen2024culturax}, Colossal OSCAR \cite{abadji-etal-2022-towards}, and HLPT v1 \cite{de-gibert-etal-2024-new}. To ensure linguistic consistency and quality, \emph{latxa-corpus-v1.1} was systematically normalized, deduplicated, and exhaustively filtered.

\begin{figure}[t]
\centering
    \includegraphics[width=0.325\linewidth]{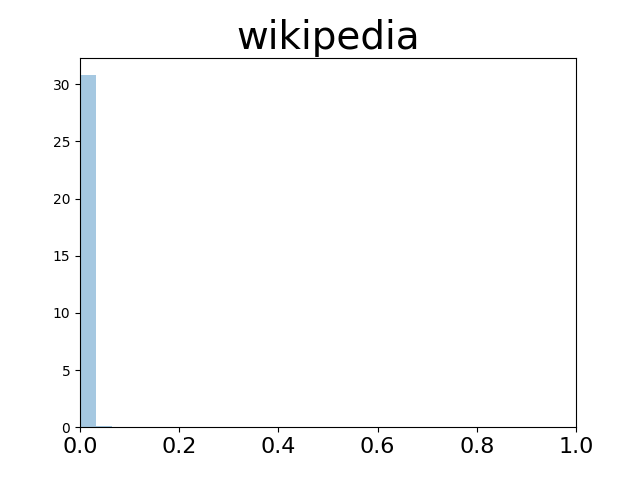} \hfil 
    \includegraphics[width=0.325\linewidth]{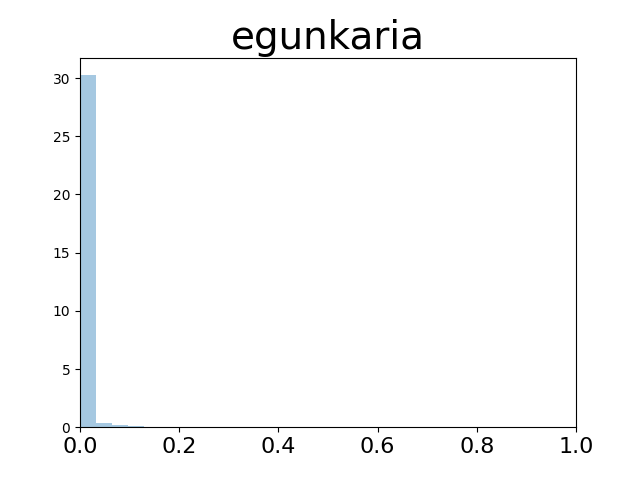}\hfil
    \includegraphics[width=0.325\linewidth]{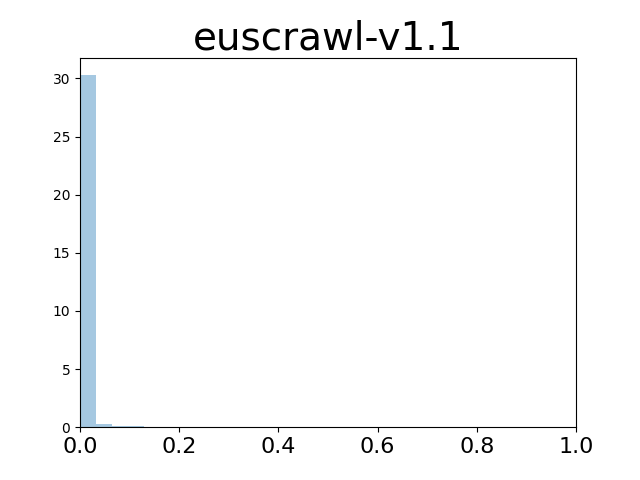}
    \includegraphics[width=0.325\linewidth]{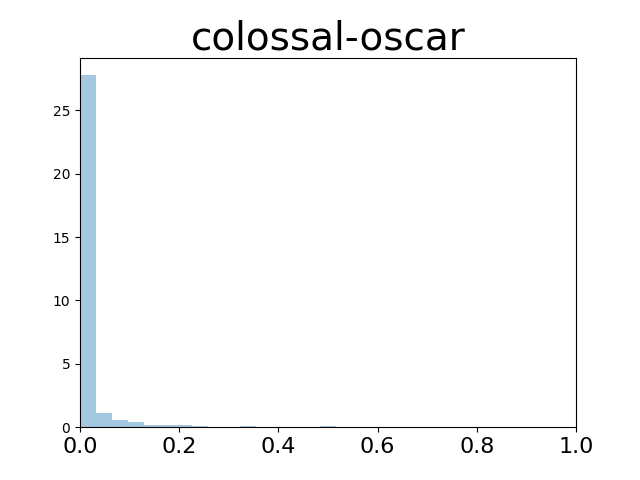} \hfil 
    \includegraphics[width=0.325\linewidth]{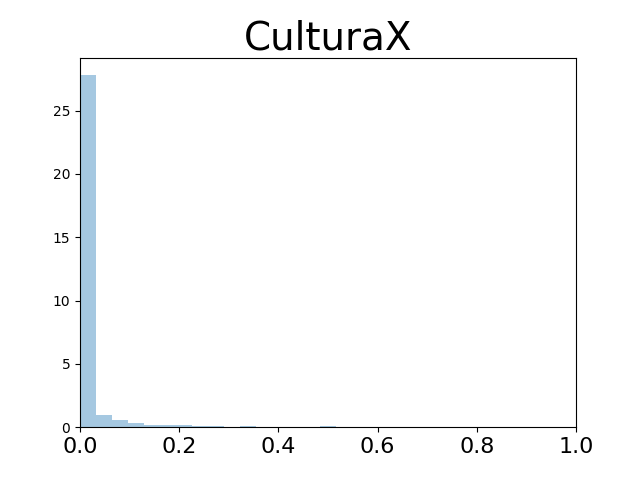}\hfil
    \includegraphics[width=0.325\linewidth]{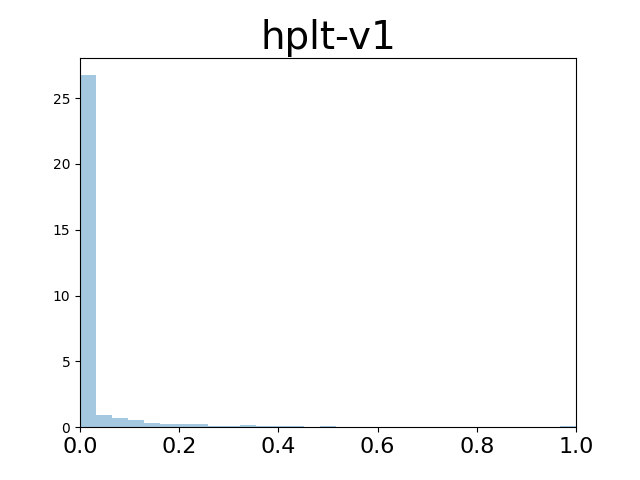}
    \includegraphics[width=0.325\linewidth]{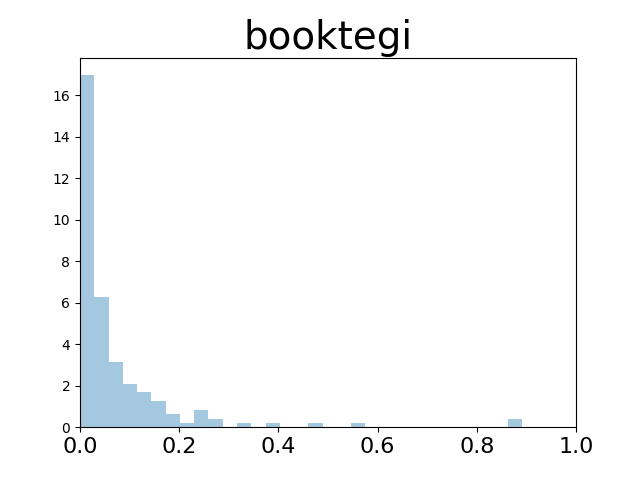} \hfil 
    \includegraphics[width=0.325\linewidth]{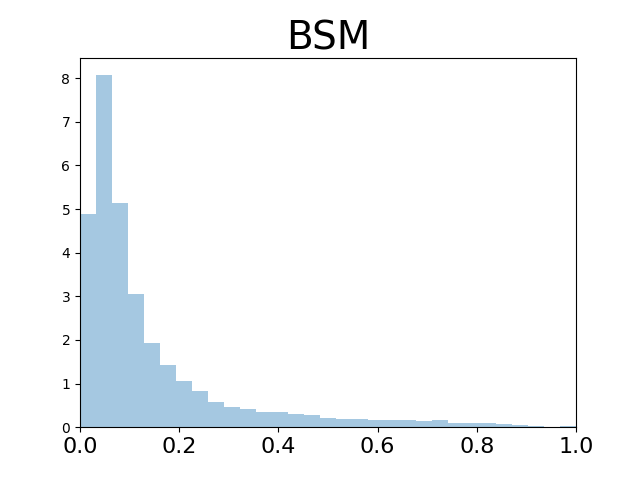}
    \includegraphics[width=0.325\linewidth]{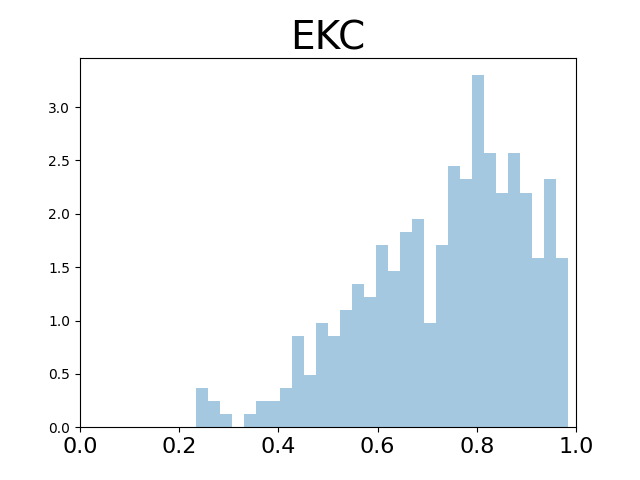}
    \caption{\footnotesize Visualization of diversity distribution for latxa standard corpora (\textit{wikipedia, egunkaria, euscrawl-v1.1, colossal-oscar, CulturaX, hplt-v1, booktegi}) and newly added \textit{BSM} and \textit{EKC} non-standard corpora.}
    \label{fig:emb_vis_eus} \hfil
\end{figure}

\subsection{Diverse Corpora}
\label{sec:diverse_corpora}

We create \emph{Diverse Corpora} by combining text from two linguistically rich and varied sources: social media and historical texts. Social media captures informal, spontaneous, and up-to-date language use, including dialectal variation, slang, and code-switching. Historical texts, in turn, provide access to pre-standard Basque, which is highly rich in dialectal diversity. \emph{Diverse Corpora} represents the largest and most diverse non-standard corpus constructed to date, with its overall size being half that of EusCrawl-v1.1 (refer to Table \ref{tab:train-data-analysis}).

\begin{table*}[t]
\centering
\scriptsize
\begin{tabular}{@{}ll rr l@{}}
\toprule
\textbf{Dataset} & \textbf{Task} & \textbf{Train/Dev} & \textbf{Test} & \textbf{Domain}\\
\midrule
BTHC\textsubscript{ 1} & Topic classification     & 10,442  & 1,854   & News \\
Korref\textsubscript{ 2} & Correference resolution  & 1,306   & 587     & News \\
NERCid\textsubscript{ 3} & NERC                   & 64,475  & 35,855  & News \\
NERCod\textsubscript{ 1} & NERC                   & 79,420  & 14,462  & News, Wikipedia \\
QNLI\textsubscript{ 1} &  QA/NLI                    & 1,994   & 238     & Wikipedia \\
WiC\textsubscript{ 1} & WSD & 409,159 & 1,400   & News \\
XNLIeu-nat\textsubscript{ 4} \dag & NLI               & 392,000      & 621     & News\\
POSud\textsubscript{ 5} \dag & POS tagging            & 7,194   & 1,799   & News, literary texts \\
POShis-nor \textcolor{red}{\bf{[new]}} \dag & POS tagging           & 5,095      & 526  & Normalized historical texts \\
\midrule
BEC\textsubscript{ 1} & Sentiment Analysis        & 7,380   & 1,302   & Twitter \\
Intent\textsubscript{ 6} & Intent classification  & 5,322   & 1,087   & Facebook \\
Slot\textsubscript{ 6}& Slot filling             & 30,443  & 5,633   & Facebook \\
Vaxx\textsubscript{ 7} & Stance detection         & 1,070   & 312     & Twitter \\
XNLIeu-var\textsubscript{ 8} \dag & NLI               & 392,000      & 894     & Dialectal language \\
POShis \textcolor{red}{\bf{[new]}} \dag & POS tagging           & 5,095      & 526  & Historical texts \\
\bottomrule
\end{tabular}
\caption{\footnotesize Dataset characteristics and sizes for Basque benchmarks. The upper block includes standard language datasets, while the lower block includes diverse language datasets (non-standard, dialectal, or historical). Datasets marked with $\dag$ are not part of BasqueGLUE. Sources: $^1$\cite{urbizu-etal-2022-basqueglue}, $^2$\cite{soraluze2012mention},$^3$\cite{alegria2006lessons},$^4$\cite{heredia-etal-2024-xnlieu},$^5$\cite{de-marneffe-etal-2021-universal},$^6$\cite{lopez-de-lacalle-etal-2020-building},$^7$\cite{agerri2021vaxxstance},$^8$\cite{bengoetxea-etal-2025-lost}.}
\label{tab:basque-dataset-sizes}
\end{table*}

\paragraph{BSM:} The Basque Social Media corpus (BSM) provides a valuable source of personal, spontaneous, and varied written content. It includes standard language as well as a wide range of dialectal, slang, informal, and code-switched data \cite{joseba-heldugazte}. Building on the same techniques and extending previously collected resources \cite{fernandez-de-landa-etal-2024-uncovering}, BSM contains approximately 11 million posts produced by more than 13,000 Basque-speaking users, amounting to around 188 million words. We perform a data augmentation step that effectively doubles the dataset size by reordering the existing material in two distinct ways. Specifically, we divide BSM into two complementary subsets containing the same textual content but differ in organization: i) BSMtime, where posts are ordered chronologically by publication time, independent of author identity (11M posts). ii) BSMauthor, where posts are grouped by author, forming 13K complete individual documents representing their timelines. This dual organization enables the analysis of both diachronic language variation and author-specific linguistic behavior, providing a flexible resource for a range of sociolinguistic and computational studies.

\paragraph{EKC:} The Corpus of Basque classical writers (\textit{Euskal Klasikoen Corpusa}, EKC) \cite{euskara2013euskal} aims to serve as the repository for nearly all classical texts up to the 20th century. It contains 338 documents or books ranging from the 16th century to 1975, comprising a total of 21 million words. This corpus covers various literary genres, including poetry, narrative, theater, essays, and religious texts. The texts originate from different dialects, as the standard Basque language was not established until the late 20th century. Due to the historical span and dialectal variety represented in the collection, the corpus is expected to be highly diverse.

\subsection{Language diversity analysis} %
\label{sec:diversity}

We conduct a language diversity analysis on the entire pre-training data to verify that the sources align with the theoretical distinction between standard and diverse language. To that end, we employ the method presented in \cite{basqueyoungtwitter}, which automatically classifies each sentence in a document as either standard or non-standard. Using these classifications, the diversity of each document is computed as the proportion of non-standard sentences relative to the total number of sentences. Subsequently, the language diversity of each source in the corpus is determined by calculating the mean diversity across all documents within that source.

\cref{tab:train-data-analysis} and \cref{fig:emb_vis_eus} present the results of our language diversity analysis across the standard corpora of \emph{latxa-corpus-v1.1} corpus as well as our newly proposed \emph{Diverse Corpora}. The results show that, on the one hand, standard language sources such as Wikipedia, Egunkaria, and EusCrawl exhibit very low levels of non-standard usage, indicating low linguistic diversity. Although CulturaX and Colossal-oscar are both large-scale multilingual corpora derived from Common Crawl, their diversity values are slightly below those of HPLT-v1, likely due to the more extensive preprocessing applied to produce cleaner, more standardized text. Booktegi, though small, contains literary genres (fiction, essays, and poetry) that are inherently diverse in syntax and vocabulary, which explains its slightly higher diversity. On the other hand, the BSM corpus, representing social media text, reaches diversity levels more than double those of the most heterogeneous standard corpus. Finally, the EKC corpus stands out with a mean diversity of $0.733$, an order of magnitude greater than any other source. This high diversity observed in the BSM and especially the EKC corpora confirms their role as essential complements to standardized resources for modeling the full spectrum of Basque linguistic variation.

\section{Evaluation datasets}
\label{sec:evaluation-datasets}

We use Natural Language Understanding (NLU) tasks to assess the effect of language diversity on Basque. Hence, we evaluate the models on all tasks from BasqueGLUE \cite{urbizu-etal-2022-basqueglue}, the standard benchmark for Basque NLU, as well as Natural Language Inference (NLI) and Part-of-Speech (POS) tagging tasks. We divide each dataset into \textit{standard} and \textit{diverse} subsets, which allows us to analyze how linguistic diversity influences model performance. Details of each dataset, including size and domain, are described in Table \ref{tab:basque-dataset-sizes}.

\paragraph{BasqueGLUE.} 
This dataset collection consists of ten tasks that cover a wide spectrum of Basque NLU challenges. The benchmark includes datasets from standard language domains such as news and literature, as well as collections drawn from more diverse contexts, particularly social media, where dialectal and non-standard Basque are widely used \cite{joseba-heldugazte}. %
The standard subset of BasqueGLUE includes topic classification (BHTC), coreference resolution (Korref), question answering framed as NLI (QNLI), Word Sense Disambiguation (WSD) in context (WiC), and named entity recognition/classification in both in-domain (NERCid) and out-of-domain (NERCod) settings. The diverse subset, in contrast, covers sentiment analysis (BEC), stance detection (Vaxx), intent classification (Intent), and slot filling (Slot).

\paragraph{Natural Language Inference.} We use the Basque NLI dataset introduced in~\cite{heredia-etal-2024-xnlieu}, which extends the XNLI benchmark \cite{conneau-etal-2018-xnli} to this language. The standard subset consists of the so-called XNLIeu-nat part of the dataset, a test set created entirely from scratch by native Basque speakers. The diverse subset consists of the NLI dataset in~\cite{bengoetxea-etal-2025-lost}, where XNLIeu-nat was adapted into three Basque dialects (Western, Central, and Navarrese), therefore allowing NLI evaluation in both standard and diverse language settings. %
For training, we rely on the automatically translated XNLIeu training set, derived from the English XNLI corpus, which naturally retains certain domain characteristics from the original English data.

\paragraph{POS tagging.} The standard subset consists of the Universal Dependencies POS (POSud) dataset \cite{de-marneffe-etal-2021-universal}, based on the Basque UD treebank \cite{aranzabe2015automatic}, which contains 8,993 sentences (121,443 tokens) drawn primarily from literary and journalistic sources. The diverse subset is POShis, a newly developed dataset targeting historical language. POShis is derived from the BIM corpus \cite{10.1093/llc/fqab066}, which comprises historical texts dating from the 15th to the mid-18th century, spanning multiple historical dialects (Western, Central, Labourdin, Souletin). We select 5,621 instances (693,414 tokens) instance from this morphosyntactically annotated diachronic corpus, making the resulting POShis subset approximately five times larger than POSud. The manual annotations have been converted to follow the same criteria as POSud. %
Additionally, we provide a parallel normalized version, POShis-nor, to facilitate evaluation under standard conditions.

\begin{table}[t]
  \centering
  \small
\begin{tabular}{l|ccc}
\toprule
                          & \multicolumn{3}{c}{\textbf{Corpora}}                 \\
Tokenizer                 & \textbf{Standard} & \textbf{Diverse} & \textbf{Both} \\
\midrule                                                                         
Tok$_{\mathrm{Standard}}$ & 1.29             & 1.62            & 1.53         \\
Tok$_{\mathrm{Diverse}}$  & 1.37             & 1.42            & 1.41         \\
Tok$_{\mathrm{Both}}$     & 1.30             & 1.45            & 1.41         \\
\bottomrule
\end{tabular}
\caption{Token fertility for different tokenizers. Rows correspond to tokenizer's training corpora, and columns to evaluation corpora (lower is better).}
\label{tab:tokens_per_word}
\end{table}
\begin{table*}[t]
\centering
\scriptsize
\begin{tabular}{ll|ccc|ccc|ccc}
\toprule
& & \multicolumn{3}{c}{\textbf{BERnaT$_{\mathrm{standard}}$}} & \multicolumn{3}{c}{\textbf{BERnaT$_{\mathrm{diverse}}$}} & \multicolumn{3}{c}{\textbf{BERnaT}} \\
& & med & base & large & med & base & large & med & base & large \\
\midrule \midrule
\multirow{9}{*}{Standard tasks}
&BHTC  & 74.85 & 75.37 & 77.53 & 73.37 & 73.97 & 76.70 & 75.06 & 75.56 & \textbf{77.83}\\
&Korref  & 64.45 & 58.66 & 58.04 & 57.13 & 59.00 & 61.61 & 60.70 & 63.54 & \textbf{69.73}\\
&NERCid  & 83.35 & 83.83 & \textbf{86.48} & 80.19 & 75.50 & 83.34 & 81.33 & 83.39 & 84.97\\
&NERCod  & 73.29 & 75.14 & 75.04 & 65.99 & 67.77 & 70.63 & 72.60 & 75.67 & \textbf{76.06}\\
&QNLI  & 69.61 & 73.25 & \textbf{74.23} & 70.73 & 70.73 & 71.01 & 69.19 & 71.15 & 72.96\\
&WiC  & 67.69 & 70.57 & 70.07 & 68.86 & 70.33 & 70.14 & 69.07 & 69.55 & \textbf{70.86}\\
&XNLIeu-nat  & 67.31 & 71.28 & \textbf{74.93} & 61.89 & 66.72 & 64.52 & 66.51 & 67.95 & 72.79\\
&POSud  & 94.85 & 95.61 & \textbf{96.27} & 94.06 & 95.42 & 95.38 & 94.46 & 95.86 & 95.97 \\
&POShis-nor  & 71.48 & 74.29 & 78.92 & 72.73 & 72.55 & 76.96 & 73.10 & 76.12 & \textbf{79.74}\\
\midrule
&\textbf{AVG} standard tasks & 74.10 & 75.33 & 76.83 & 71.66 & 72.44 & 74.48 & 73.56 & 75.42 & \textbf{77.88} \\
\midrule \midrule
\multirow{7}{*}{Diverse tasks}
&BEC  & 69.87 & 66.05 & 68.10 & 69.43 & 69.82 & \textbf{70.10} & 68.87 & 70.02 & 69.40\\
&Intent  & 75.31 & 77.46 & 77.80 & 77.34 & 75.50 & \textbf{79.70} & 75.22 & 76.84 & 78.04\\
&Slot  & 76.73 & \textbf{78.95} & 77.01 & 76.93 & 78.26 & 76.01 & 78.02 & 75.75 & 78.61\\
&Vaxx  & 61.60 & 63.06 & 65.97 & 64.03 & 66.67 & \textbf{68.46} & 66.13 & 65.77 & 67.44\\
&XNLIeu-var & 65.21 & 68.05 & \textbf{74.05} & 58.50 & 65.14 & 63.68 & 62.30 & 64.91 & 72.82\\
&POShis  & 73.06 & 73.96 & 75.84 & 73.22 & 73.17 & 73.27 & 73.00 & 74.36 & \textbf{76.33}\\
\midrule
&\textbf{AVG} diverse tasks & 70.30 & 71.26 & 73.13 & 
69.91 & 71.43 & 71.87 & 
70.59 & 71.28 & \textbf{73.77} \\
\midrule \midrule
\textbf{AVG} overall && 72.58 & 73.70 & 75.35 & 
70.96 & 72.04 & 73.43 & 
72.37 & 73.76 & \textbf{76.24} \\
\bottomrule
\end{tabular}%
\caption{Results by task for all BERnaT model variants and sizes, divided by standard and diverse tasks.}
\label{tab:results_main}
\end{table*}
\section{Training Models}
The main aim of the experiments is to evaluate the effect of including diverse language data in pre-training on NLU tasks. To that end, we start by pre-training the models with the corpora described in Section \ref{sec:corpora}, which are then finetuned to each task and dataset (c.f. Section \ref{sec:evaluation-datasets}). In this section, we describe these steps in turn.

\paragraph{Pretraining.} 
We use encoder-only models that follow the RoBERTa \cite{liu2019roberta} architecture for several model sizes: medium (51M), base (124M) and large (355M). We train the models from scratch, using three corpora combinations (standard, diverse and both) therefore generating BERnaT$_{\mathrm{standard}}$, BERnaT$_{\mathrm{diverse}}$ and BERnaT models, respectively (see Figure \ref{fig:model_summary}). We follow \cite{artetxe-etal-2022-corpus-robertaeuscrawl} and pre-train each model with 1 million tokens with a sequence length of 512 tokens. We use packing, i.e., shorter documents are concatenated or packed into a single sequence to fully utilize the sequence length and maximize the efficiency of each training iteration. Documents that exceed the sequence length, particularly those from Booktegi and EKC, are divided into smaller chunks. We train the models up to 100 epochs and choose the best checkpoint based on the validation loss. To speed up pre-training, we use Flash-Attention 2 \cite{dao2023flashattention2fasterattentionbetter} and mixed-precision (FP16), as preliminary experiments have shown no difference in performance compared to full precision and regular attention implementation.

\paragraph{Handling diversity.}
During the initial phase of training the standard models, we adopted the hyperparameter configuration proposed by \cite{artetxe-etal-2022-corpus-robertaeuscrawl}. Applying the same configuration to the more heterogeneous corpora, however, frequently yielded exploding gradients. This instability is likely due to the increased data diversity, and the corresponding sensitivity of the model under such a distribution. To mitigate this issue, we progressively reduced the learning rate until stable convergence was achieved. Preliminary experiments showed a clear correlation between corpus diversity and the learning rate required for stability: as diversity increases, smaller learning rates are needed to prevent exploding gradients. Accordingly, we identified learning rates of  $8e^{-4}$, $4e^{-4}$ and $1e^{-4}$ as good candidates for stable training of medium, base, and large models, respectively. We maintain these learning rates across all BERnaT variants to enable fair comparison. For the remaining hyperparameters, we follow the configuration proposed by \cite{artetxe-etal-2022-corpus-robertaeuscrawl}.
\begin{figure*}[t]
\centering
\includegraphics[scale=0.43]{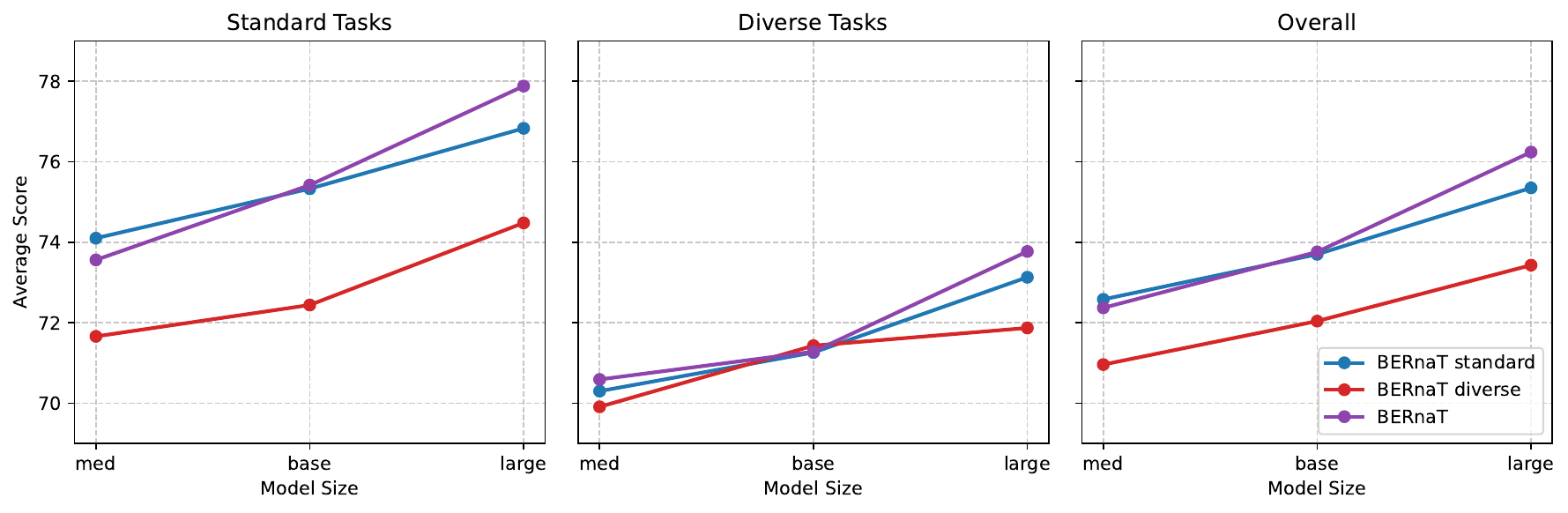}
\caption{\footnotesize Average performance of Diverse, Standard and combined models in three sizes on Standard, Diverse and all tasks.}\label{fig:bernat_eval}
\end{figure*}
\paragraph{Tokenizers.} 

To accurately represent the data on our corpora, and given that the target language is morphologically rich and exhibits substantial cross-domain lexical variation, we train a specialized BPE tokenizer for each corpus combination, with a vocabulary size of 50k \cite{liu2019roberta}. \cref{tab:tokens_per_word} reports the token fertility~\cite{rust-etal-2021-good} of each tokenizer (denoted Tok$_{\mathrm{Standard}}$, Tok$_{\mathrm{Diverse}}$, and Tok$_{\mathrm{Both}}$) across all corpus configurations. Token fertility measures the average number of subword tokens required to represent a single word, providing insight into the tokenizer’s efficiency and its alignment with the linguistic properties of the data. Lower fertility values indicate more compact tokenization, reflecting a closer match between the tokenizer’s learned vocabulary and the underlying lexical structure of the corpora. \cref{tab:tokens_per_word} shows that tokenizers trained on specific corpora obtain the lowest token fertility on that corpora, as expected, with the tokenizer trained with both standard and diverse corpora obtaining the second-best token fertility values.

\paragraph{Fine-tuning.} For the tasks included in BasqueGLUE, we use the same training configuration and hyperparameters as in \cite{artetxe-etal-2022-corpus-robertaeuscrawl,urbizu-etal-2022-basqueglue}, except for the large models, where we halve the learning rate to reduce excessive variation across random seeds. For POSud, POShis-nor, and POShis, we use their respective training sets for training and evaluate on the corresponding test sets. All POS tasks use the same hyperparameters as those employed to fine-tune the NERC tasks in BasqueGLUE. For the NLI tasks, we use XNLIeu-nat and XNLIeu-var as test sets, while training on the XNLIeu training set, following the same hyperparameters as in \cite{bengoetxea-etal-2025-lost}. We perform three runs for each task and report the average performance along with the variation across runs.

\section{Results}

Table \ref{tab:results_main} summarizes the performance of all BERnaT model variants (BERnaT$_{\mathrm{standard}}$, BERnaT$_{\mathrm{diverse}}$, and BERnaT) on the proposed set of standard and diverse NLU tasks. Models are evaluated in three sizes (medium, base, and large), enabling analysis of the effects of both training data composition and model scale. %

\paragraph{Including diverse corpora is beneficial.} 
The BERnaT models trained with both standard and diverse corpora achieve the best results overall on both standard and diverse datasets. This shows that including non-standard text in pretraining improves model performance not only on NLU tasks involving informal, dialectal, or historical texts, but also on tasks dealing with standard Basque texts. 

In the standard subset (Table \ref{tab:results_main}, upper half), BERnaT$_{\mathrm{standard}}$ unsurprisingly achieves strong results. However, on average, these results are still below those achieved by the combined BERnaT model. While standard models perform well on tasks such as NERCid, QNLI, XNLIeu-nat, and POSud, the combined model achieves higher average performance across all large-model tasks. This indicates that exposure to diverse data does not harm performance on conventional benchmarks; on the contrary, the mixed-corpus BERnaT retains or improves accuracy on most tasks. 
\begin{figure*}[ht]
\centering
\includegraphics[width=\linewidth]{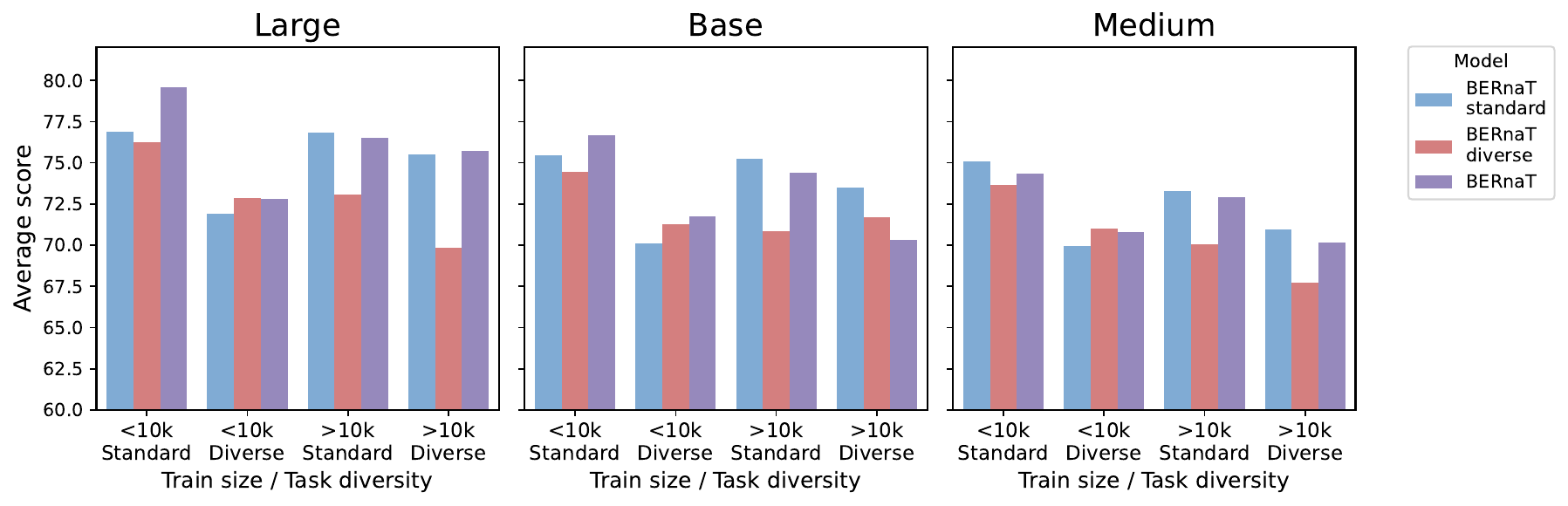}
\caption{Average performance of Diverse, Standard and combined models, with results grouped according to training size and diversity of the task as presented in Table~\ref{tab:basque-dataset-sizes} (standard/diverse). Each plot corresponds to the size of the models (medium/base/large). }\label{fig:average-fine-tuning-size}
\end{figure*}
The diverse subset (Table \ref{tab:results_main}, lower half) highlights the advantage of training with linguistically heterogeneous data. The BERnaT$_{\mathrm{diverse}}$ models surpass their standard counterparts on several tasks, including Intent, Vaxx, and BEC, showing a clear correlation between pre-training data composition and task-specific performance, except for the NLI tasks, where BERnaT$_{\mathrm{diverse}}$ yields results up to 13 points lower than  BERnaT$_{\mathrm{standard}}$, affecting its average scores. This stark contrast appears in both XNLIeu-nat and XNLIeu-var, which suggests that it is not a consequence of the diversity of the tasks, but rather of the task itself. Nonetheless, the combined BERnaT again delivers the best average performance on diverse tasks, slightly above both the standard and diverse variants, confirming that mixing standard and diverse sources leads to a more balanced and generalizable representation space. All in all, these results show that data diversity is particularly beneficial when balanced with standard language data: overly diverse corpora may introduce stylistic or lexical noise, but when paired with curated standard text, they help models capture a wider linguistic spectrum (variation across dialects, registers, social groups or even time) without losing precision on standard tasks.%

\paragraph{Model size matters.} 
The increase in performance resulting from including diverse corpora in pretraining varies according to the model size, as shown in Figure \ref{fig:bernat_eval}. Smaller models (med) do not benefit when including diverse text, and for most tasks (6 out of 9 standard tasks and 4 out of 6 diverse tasks), it is more effective to train specialized models using either standard or non-standard texts. With base models, using both standard and diverse corpora slightly underperforms compared to specialized models. As previously mentioned, large models gain the most from the inclusion of the entire corpus. Moreover, the performance boost increases with model size, not only on diverse tasks but also on standard benchmarks. %

\paragraph{The impact of fine-tuning data size.} We analyze whether the benefits of including diverse data in pre-training diminish with the size of the annotated data used to finetune the models for downstream tasks. In Figure~\ref{fig:average-fine-tuning-size}, the evaluation tasks are grouped into two main categories, depending on the size of the training split: those with fewer than 10K instances and those with more than 10K instances. While minor differences can be found depending on the model size, a distinct trend can also be perceived. For models trained on smaller training sets,  BERnaT$_{\mathrm{standard}}$ outperforms BERnaT$_{\mathrm{diverse}}$ in standard tasks, whereas the opposite holds for diverse tasks. This behavior aligns with our expectations: when the pre-training and the fine-tuning data match in diversity, the performance of the model improves. Nevertheless, with an increase in the size of fine-tuning data, this match is no longer necessary, as we can see that BERnaT$_{\mathrm{standard}}$ achieves the best results in almost all cases, regardless of diversity. The gains from combining standard and diverse data during pre-training are therefore most notable when fine-tuning data are scarce, which suggests that the smaller the training dataset for fine-tuning, the more important the diversity of models' pre-training models becomes for downstream performance. In any case, the full BERnaT is a good compromise that obtains the best results on average.

\begin{table}[t]
    \centering
    \resizebox{\columnwidth}{!}{
    \begin{tabular}{lccc}\toprule
         &  \textbf{central}&  \textbf{western}& \textbf{navarrese} \\\midrule
         BERnaT&  75.13&  72.08& 66.67 \\
         BERnaT$_{\mathrm{standard}}$&  \textbf{75.47}&  \textbf{72.50}& \textbf{69.84} \\
         BERnaT$_{\mathrm{diverse}}$&  64.64&  58.75& 61.90 \\ \bottomrule
    \end{tabular}
    }
    \caption{Accuracy per dialect obtained by the large models of the BERnaT family.}
    \label{tab:analysis-dialect}
\end{table}

\paragraph{NLI Dialect Results.} Finally, we analyze NLI performance of BERnat models for different Basque dialects. Table~\ref{tab:analysis-dialect} shows the results of the large BERnaT models on three dialects, central, western, and navarrese. Similar to the results by \cite{bengoetxea-etal-2025-lost}, all BERnaT models show a clear preference for the central variety, as it is closer to standard Basque, followed by western and navarrese. Surprisingly, BERnaT$_{\mathrm{standard}}$ shows slightly higher results than BERnaT, and BERnaT$_{\mathrm{diverse}}$ lags behind both by a large margin. We attribute these results to the effect of the size of fine-tuning data as discussed earlier, given the considerable size of the training data used for the XNLIeu-var task.

\section{Conclusion}

This study examines the impact of linguistic diversity on language model pre-training through the lens of Basque, a low-resource language. We introduce the BERnaT family of encoder-only models, each trained on different types of corpora: standard, diverse, and combined. To assess their performance, we introduce a novel evaluation framework comprising a comprehensive suite of Natural Language Understanding (NLU) tasks, organized into two distinct subsets: standard and diverse evaluation benchmarks.

Our findings demonstrate that combining standard and diverse data yields the best overall performance. Models trained only on diverse data do not outperform standard models, but the integration of both leads to substantial gains in generalization across linguistic contexts. Importantly, exposure to diverse linguistic data does not degrade performance on standard benchmarks, indicating that diversity and quality are not mutually exclusive.

We also observe that larger models benefit more from data diversity, suggesting that increased capacity enhances the model’s ability to internalize complex linguistic variation. Furthermore, tasks involving social media or historical texts benefit most from diverse pre-training, confirming the practical value of incorporating diverse language data in low-resource settings.

In conclusion, our results underscore the importance of linguistic diversity in building equitable and robust language technologies. The balanced inclusion of standard and clean texts with informal, dialectal, and pre-standard language variations is key not only for fair representation but also for improved performance and generalization in multilingual and low-resource contexts.

\section*{Limitations} 
Despite the advances presented in our paper, our approach remains constrained by the availability of labeled data for diverse language tasks and the limitations of encoder-only architectures. We will therefore explore the use of both larger model scales and generative architectures in few-shot or zero-shot settings. This direction is promising for capturing complex linguistic variation with limited labeled training data. We will also integrate generation-based evaluations to assess linguistic adaptability beyond fine-tuning.

\nocite{*}

\bibliographystyle{fullname}
\bibliography{EjemploARTsepln}

\end{document}